\documentclass[11pt,a4paper]{article}
\usepackage[hyperref]{ranlp2023}
\usepackage{times}
\usepackage{latexsym}

\usepackage{microtype}

\aclfinalcopy 

\usepackage{amssymb}
\usepackage{amstext}
\usepackage{amsmath}
\usepackage{amsthm}
\usepackage{bbold}
\usepackage{graphicx}
\usepackage{float}
\usepackage{enumitem}
\usepackage{mathtools, nccmath}
\usepackage{hyperref}

\title{Non-Parametric Memory Guidance for \\ Multi-Document Summarization}

\author{Florian Baud \\
  LIRIS, Villeurbanne \\
  Visiativ \\
  \texttt{florian.baud@visiativ.com} \\\And
  Alex Aussem \\
  LIRIS, Villeurbanne \\
  \texttt{alexandre.aussem@liris.cnrs.fr} \\}

\date{}

\begin{document}

\maketitle
\begin{abstract}
Multi-document summarization (MDS) is a difficult task in Natural Language Processing, aiming to summarize information from several documents. However, the source documents are often insufficient to obtain a qualitative summary. We propose a retriever-guided model combined with non-parametric memory for summary generation. This model retrieves relevant candidates from a database and then generates the summary considering the candidates with a copy mechanism and the source documents. The retriever is implemented with Approximate Nearest Neighbor Search (ANN) to search large databases. Our method is evaluated on the \textit{MultiXScience} dataset which includes scientific articles. Finally, we discuss our results and possible directions for future work.
\end{abstract}

\section{Introduction}

Multi-document summarization is performed using two methods: extractive \cite{wang-etal-2020-heterogeneous, liu-etal-2021-hetformer} or abstractive \cite{jin-etal-2020-multi, xiao-etal-2022-primera}. So-called extractive methods rank sentences from source documents that best summarize them. These methods reuse important information well to construct a good summary but they lack coherence between sentences. To overcome this issue, abstractive methods are studied to imitate human writing behavior. They show great performance in human writing style but they often miss key information.


To make abstractive models aware of essential information, \cite{dou-etal-2021-gsum} guides their model with additional information like a set of keywords, graph triples, highlighted sentences of source documents, or retrieved similar summaries. Their method, which uses every guidance previously mentioned, improves summary quality and controllability compared with unguided models. However, guidances require specific training data, especially for keywords, graph triples, and highlighted sentences.


Our proposal is that by guiding with pre-existing summaries, the model can draw inspiration from the summary as a whole. But also be able to extract keywords and phrases using a copy mechanism. Consequently, this work focuses on guidance by similar summaries extracted from a knowledge base using a similarity metric between source documents and pre-existing summaries. The model, inspired by RAG \cite{10.5555/3495724.3496517}, is fully differentiable. In addition, the model generator uses a copy mechanism on the candidates returned from the knowledge base, inspired by \cite{cai-etal-2021-neural}. The findings of these two studies motivated the development of our model for the multi-document text summarization task.

We demonstrate the potential of our method on \textit{MultiXScience} \cite{lu-etal-2020-multi-xscience}. This dataset gathers scientific articles where we have to generate the \textit{"related work"} part with the \textit{"abstract"} of the source article and the \textit{"abstracts"} of the citations. In the case of scientific articles, we believe that the source documents are insufficient to generate the \textit{"related work"} part because external knowledge is necessary to write such a paragraph.

\begin{figure*}[h]
    \begin{center}
    \includegraphics[scale=0.75]{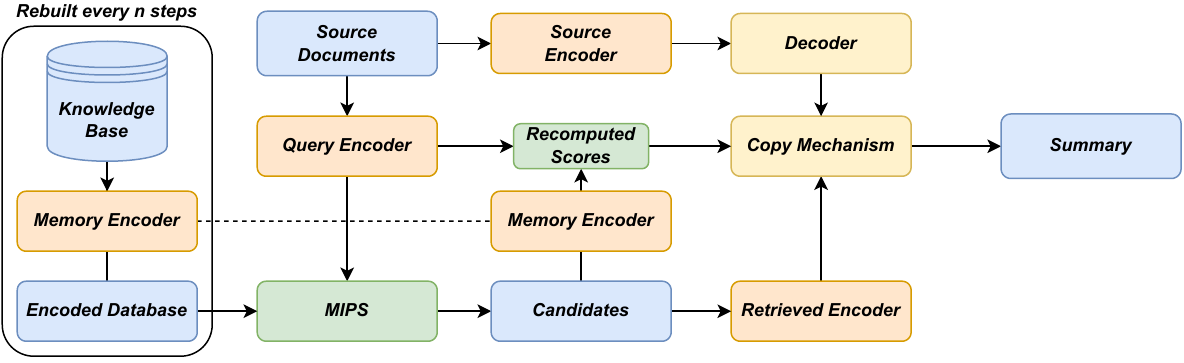}
    \end{center}
    \caption{In the first step, the knowledge base is built by encoding all documents with the memory encoder. Then the source documents are transformed with a query encoder and with a source encoder, the query encoder is used to search the knowledge base. The encoded source is used to represent the source documents for the generation of the summary. After retrieving the top-$k$ of the search, they are encoded with the retrieved encoder and again with the memory encoder to recalculate the relevance score for back-propagation. Then, the decoder takes as input the source documents and the relevant documents for the generation of the summary.}
    \label{fig:model}
\end{figure*}



In this work, we investigate a sequence-to-sequence model guided by a memory retriever of similar summaries. Specifically, source documents are the input of the memory retriever, which returns the top k similar summaries from a potentially large database using an approximate nearest neighbor search. Then, the decoder generates the summary taking into account the source and retrieved summaries and is trained to identify interesting texts for the targeted summary. The code of our work is available on GitHub\footnote{\href{https://github.com/florianbaud/retrieval-augmented-mds}{https://github.com/florianbaud/retrieval-augmented-mds} (visited on 11/08/2023)}.

Our contribution is twofold: firstly, we integrate a retriever to retrieve candidates for the generation of the summary, and secondly, we make use of a copy mechanism to incorporate these candidates into the generation procedure.




\section{Related Work}


We start with a brief review of related work.
\cite{cohan-etal-2018-discourse} proposes to capture the structure of the document to better represent the information of the source document. Their method is applied to scientific articles from \textit{Arvix} and \textit{Pubmed} which are long documents. For the same purpose, \cite{Cohan2018, Yasunaga_Kasai_Zhang_Fabbri_Li_Friedman_Radev_2019} propose to generate a summary from the articles that cite the article to be summarised. The disadvantage of these methods is that they cannot be used when writing an article. In this work, we use references and not the papers that cite the documents to be summarised. More recently, \cite{xiao-etal-2022-primera} proposed a pre-training strategy dedicated to multi-document text summarisation, their masking strategy showed significant improvement for the MDS task. They applied their method to the \textit{MultiXScience} dataset.


The models using guidances are close to our work, indeed \cite{cao-etal-2018-retrieve, dou-etal-2021-gsum} use retrieved summaries to better control the summary generation. However, they use information retrieval systems such as \textit{ElasticSearch} to find candidates for summary generation. Also, \cite{an2021retrievalsum} has introduced dense search systems for text summarization, but they do not train the retriever with the summary generator. In our case, the retriever is dense and trainable to find the most relevant candidates for the generation of the summary.

In addition, retrieval-augmented models share commonalities with our work. RAG, \cite{10.5555/3495724.3496517} which introduced this type of model, is used for the question-answering task, where a context is given to answer the question. The model retrieves several contexts with a retriever and then answers the question using each of the retrieved candidates. These types of models are also used in the translation task, where \cite{cai-etal-2021-neural} translates a sentence with a pre-established translation base. Their model searches this base for translations close to the sentence to be translated and then incorporates them into the generation of the translation through a copy mechanism. This approach shares some similar intuition with our proposed approach because our architecture is based on an augmented retriever that incorporates the memory by means of a copy mechanism. It is interesting to investigate whether the encouraging success of the copy mechanism recently obtained in translation carries over to the MDS task.

\section{Proposed Method}\label{sec:model}

Inspired by \cite{cai-etal-2021-neural}, we propose a model composed of a memory retriever and a copy generator. Figure \ref{fig:model} illustrates our framework, where we start by encoding the entire knowledge base. After an arbitrary number of steps during the training, the encoded knowledge base is updated. Then, the forward pass encodes source documents and finds similar documents. Retrieved documents are encoded and fed to the generator with the source documents. Our memory retriever has multiple encoders, one for encoding the query, one for the knowledge base, one for the sources documents, and one for the retrieved candidates. Our copy generator is a decoder with a cross-attention mechanism on source document embeddings and a copy mechanism on retrieved candidates, which is placed at the top of the decoder. We begin describing the retriever and then show how our generator works.




\subsection{Memory Retriever}
\label{sec:memory-retriever}

The retrieval approach consists of source documents as query and documents from a knowledge base denoted respectively by $q$ and $c$. Documents are often too long to be encoded with a \textit{Transformer} \cite{NIPS2017_3f5ee243}, so we used a \textit{LongFormer} \cite{Beltagy2020Longformer} model.
\textit{LongFormer} has a Transformer-like architecture that can deal with long input sequences by attending tokens with windowed attention and global attention on a few tokens. We encode source documents and candidates documents with a pretrained \textit{LongFormer} model separated by a special token (\textit{[DOC]}) :

\begin{align*}
    h^{q} &= LED^q_{enc}(q) \\
    h^{m} &= LED^m_{enc}(m)
\end{align*}

\noindent where the \textit{LongFormer} encoder is denoted  by $LED_{enc}$. All documents in the knowledge base are encoded and stored in an index. For retrieving candidates, we take the \textit{[CLS]} token of encoders output that we normalize and we define a relevance function :

\begin{gather*}
    h^{q}_{cls} = norm(h^{q}_{cls}) \\
    h^{m}_{cls} = norm(h^{m}_{cls}) \\
    score(x, y) = x^\top \cdot y
\end{gather*}


We then calculate the relevance score on normalized tokens, which represents the cosine similarity between source documents $q$ and candidate documents $m$ that fall in the interval $[-1, 1]$. 

For fast retrieval, we retrieve the top-$k$ candidates $m_{topk} = (m_1, \ldots, m_k)$ using the maximum inner product search (MIPS) implemented with FAISS \cite{8733051}. At each training step, we calculate the actual embedding of candidates $\{h^{m}_{cls,i}\}^k_{i=1}$ and compute their relevance scores $\{s_i = score(h^m_{cls,i}, h^q_{cls})\}^k_{i = 1}$ for back-propagation as in \cite{cai-etal-2021-neural, 10.5555/3495724.3496517}. The recalculated score biases the decoder copy mechanism , which we detail in section \ref{sec:decoder}.

The memory encoder does not re-encode all the knowledge base at each training step because this would be expensive computation. Instead, the knowledge base and the MIPS index are updated at regular intervals defined arbitrarily. On the other hand, we encode the retrieved top-$k$ candidates and the source documents with two encoders, $LED^r_{enc}$ and $LED^s_{enc}$, as shown below:

\begin{gather*}
    h^{s} = LED^s_{enc}(q) \\
    h^{r}_{topk} = LED^r_{enc}(m_{topk})
\end{gather*}

These two results are forwarded to the copy generator, which we detail in the next section.

\subsection{Copy Generator}\label{sec:decoder}

In the generation part of our model, we use the decoder from \textit{LongFormer} and apply a copy mechanism to previously retrieved candidates. Formally, we have :

\begin{align*}
    h^{d} = LED_{dec}(y, h^s)
\end{align*}

\noindent where $LED_{dec}$ corresponds to the decoder part of the \textit{LongFormer} model, and $y$ is the targeted summary. The decoder attends over source documents $h^s$ and previous tokens $y_{1:t-1}$, producing a hidden state $h^d_t$ at each time step $t$. The probability of the next token is calculated with a $softmax$ function:

\begin{align}
    \label{eq:prob1}
    P_{dec}(y_t) = softmax(W_{d} \cdot h_t^d  + b_{d})
\end{align}

\noindent where $W_{d}$ is a $hiddens_{size} \times vocab_{size}$ matrix and $b_{d}$ is the bias; both are trainable parameters. 

Then, we incorporate the top-$k$ candidates $m_{topk}$ with a copy mechanism by calculating a cross attention between $h^{d}_t$ and $h^{r}_{topk}$. To this end, we reuse the cross-attention part of \textit{LongFormer} to add it after its original decoder. This new layer has only one attention head in order to use the attention weights as the probability to copy a word from top-$k$ candidates. 

Given $k$ documents encoded in $h^{r}_{topk}$, then we can construct a set of token embedding $\{r_{i,j}\}^{L_i}_{j=1}$ where $i \in [1, k]$, $j \in [1, L_i]$ and $L_i$ is the length of document $i$. Formally, the attention weight of the $j$th token in the $i$th relevant document is expressed as,

\begin{gather*}
    \alpha_{ij} = \frac{
    \exp(h^{d\top}_t W_a r_{i,j} + \beta s_i)
    }{
    \sum^k_{i=1} \sum^{L_i}_{j=1} \exp(h^{d\top}_t W_a r_{i,j} + \beta s_i)
    } \\
    c_t = W_c \sum^k_{i=1} \sum^{L_i}_{j=1} \alpha_{ij} r_{i,j}
\end{gather*}

\noindent where $\alpha_{ij}$ is the attention weight of the $j$th token in the $i$th relevant document, $W_a$ and $W_c$ are learnable parameters, $c_t$ is a weighted representation of top-$k$ candidates and $\beta$ is a learnable scalar that controls the relevance score between the retrieved candidates and the decoder hidden state, enabling the gradient flow to the candidates encoders as in \cite{cai-etal-2021-neural,10.5555/3495724.3496517}. Equation \ref{eq:prob1} may be rewritten to include the memory:

\begin{align}
    P_{dec}(y_t) = softmax(W_{d} \cdot (h_t^d + c_t)  + b_{d})
\end{align}

Thus the next token probability takes into account the attention weights of the top-$k$ candidates. The final next token probability is given by:

\begin{align*}
    P(y_t) = (1 - \lambda_t)P_{dec}(y_t) + \lambda_t \sum^k_{i=1} \sum^{L_i}_{j=1} \alpha_{ij} \mathbb{1}_{r_{ij} = y_t}
\end{align*}

\noindent where $\lambda_t$ is a gating scalar computed by a feed-forward network $\lambda_t = g(h^d, c_t)$. The model is trained with the log-likelihood loss $\mathcal{L} = -\log P(y^*)$ where $y^*$ is the target summary.


\subsection{Training Details}

Our model is composed of several encoders and one decoder based on the \textit{LongFormer} \cite{Beltagy2020Longformer} large model. Therefore, the size of our model attains $1.9$B of trainable parameters. Then we used the  \textit{DeepSpeed} \cite{10.1145/3394486.3406703} library for the training. Our model uses the \textit{LongFormer} pretrained models available on \textit{HuggingFace}\footnote{\href{https://huggingface.co/allenai}{https://huggingface.co/allenai} (visited on 11/08/2023)}.


The training of the model makes use of \textit{MultiXScience} data comprising 30,369 scientific articles for training, 5,066 validation, and 5,093 test articles. The objective is to generate the related work using the abstract of the article and the abstracts of the cited articles. This is an interesting dataset to experiment with because writing a related work part requires knowledge beyond the scope of the source documents.


\paragraph{Cold start problem}

At the beginning of the training, the weights are randomly initialized. Therefore the retriever selects low-quality candidates that don't send out a good signal for training. Under these conditions, the retriever cannot improve, and the model will ignore the retriever's candidates.
To overcome this cold start problem, we pre-trained the retriever on the \textit{MultiXScience} data to improve the quality of the retriever. The objective is to maximize the similarity between the abstract and the related work section. These two sections are encoded with the two encoders of the retriever to calculate the cosine similarity. 


In concrete terms, pre-training works as follows. For a batch size equal to $N$, we have $N$ "abstract" sections encoded with $A = \{LED_{enc}^q(a_i)\}_{i=1}^N$ and $N$ "related work" sections encoded with $B = \{ LED^m_{enc}(b_j)\}_{j=1}^N$, in order to obtain a cosine similarity equal to 1 when $j=i$ corresponds to positive examples and -1 otherwise for negative examples. We calculate for each element in $A$, the following errors:


\begin{align*}
    \mathcal{L}_i(A, B) = -\log \frac{\exp{(score(A_i, B_i) / \tau)}}{\sum^N_{j=1} \exp{(score(A_i, B_j)/ \tau)}}
\end{align*}

\noindent where $\tau$ is an arbitrarily chosen temperature parameter. The final error is $\mathcal{L} = \sum^N_{i=1} \mathcal{L}_i$ back-propagated in the two encoders of the retriever.



\section{Experiments}


In this section, we report on the experiments performed on the \textit{MultiXScience} dataset to evaluate our model. Training the full model is more difficult due to its size but also due to the cold start problem. The latter corresponds to the fact that the similar summaries retrieved are not sufficiently relevant to help the model. In addition, we have trained two other methods adapted to text summarisation as a comparison, \textit{Bart} \cite{DBLP:journals/corr/abs-1910-13461} and \textit{T5} \cite{JMLR:v21:20-074}. We detail the training procedure for each of them. All models use the beam search method to generate summaries. We chose a beam size of 4, a length penalty of 1.0, and limited the repetition of tri-grams. The rouge scores \cite{lin-2004-rouge} on the \textit{MultiXScience} dataset are reported in table \ref{rouge-score}.


\begin{table}[h]
\centering
    \begin{tabular}{llll}
    \hline
    \textbf{Method} & \textbf{R-1} & \textbf{R-2} & \textbf{R-L} \\
    \hline
    Ours & 30.6 & 6.5 & 17.7 \\
    Bart (Our run) & 32.4 & 7.2 & 17.3 \\
    T5 (Our run) & 29.6 & 6.3 & 17.0 \\
    Primera* & 31.9 & \textbf{7.4} & 18.0 \\
    PointerGenerator* & \textbf{33.9} & 6.8 & \textbf{18.2} \\
    \hline
    \end{tabular}
\caption{\label{rouge-score} The ROUGE score (R-1/R-2/R-L) of our preliminary results on the \textit{MultiXScience} test dataset. The * symbol means that the results have been borrowed from \cite{xiao-etal-2022-primera}.}
\end{table}

\paragraph{Reduced model} To reduce the computational burden, we used a reduced model where the knowledge base is not reconstructed. In addition, the memory encoder parameters were frozen in order to reduce the complexity of the training. These two modifications reduced the training time considerably. Indeed, the burden of reconstructing the knowledge base was overwhelming. The reduced model has fewer trainable parameters (1.4B). The model was trained for 12.000 steps on four v100 GPUs with Adam optimizer and a learning rate of $3e-5$, a batch size of 64, a top-$k$ of 5 for the retriever, and with 2.000 warmup steps and linear decay. Despite its reduction in size, we observe that the model is competitive with the state of the art.


\paragraph{Bart} We fine-tuned a \textit{Bart-large} model on the \textit{MultiXScience} dataset using a single v100 GPU over two days. The model weights were updated for 20,000 steps with a learning rate of 3.0e-5. A linear warmup for 2,000 steps was applied to the learning rate. We also limited the norm of the gradient to 0.1. The training aims to minimize cross-entropy with a smoothing label of 0.1. The \textit{MultiXScience} articles have been concatenated using the '\textbackslash n\textbackslash n' separator. The results show that Bart is competitive with the state of the art.


\paragraph{T5} The \textit{T5-large} model was fine-tuned on the same dataset as before. The training lasted 4 days on a single v100 GPU, this model is slightly larger and was trained in fp32 precision. As T5 is a text-to-text model, we have used the prefix 'summarize:' for the input documents, which are separated by the separator '\textbackslash n\textbackslash n'. The model was trained for 7,000 steps with a learning rate of 1.0e-4 and a batch size of 64. A linear warm-up of up to 2000 steps and a gradient norm limitation of 0.1 was applied. The error to be minimized is the cross-entropy with a label smoothing of 0.1.

\section{Conclusion and Future Work}

This paper presents an architecture for multi-document text summarization inspired by retrieval-augmented models. This architecture includes a retriever that searches a knowledge base to find relevant documents for the generation of a summary. These documents are integrated in the generation by means of a copy mechanism. A reduced version of the model was evaluated on the \textit{MultiXScience} dataset. The preliminary results are already competitive with the state of the art however we expect to improve our results further by: 1) properly fixing the cold start problem, and 2) training the full model. In the future, we also plan to increase the size of the knowledge base with new data and apply our method to other MDS benchmark datasets.

\section*{Acknowledgments}

We gratefully acknowledge support from the CNRS/IN2P3 Computing Center (Lyon - France) for providing computing and data-processing resources needed for this work. In addition, This work was granted access to the HPC resources of IDRIS under the allocation 2022-AD011013300 made by GENCI. Finally, we would like to thank Roch Auburtin from Visiativ for his advice.

\bibliographystyle{acl_natbib}
\bibliography{anthology,ranlp2023}


\end{document}